# A Novel Collision Detection and Avoidance system for Midvehicle using Offset-based Curvilinear Motion

N.Prabhakaran[1] P.M.Balasubramaniam[2*], S.Sheik Mohammed[3] P.RanjithKumar[4]

[1,2] Electrical and Electronics Engineering, Sri Shakthi Institute of Engineering and Technology, Coimbatore-62,
[3] Electrical and Electronics Engineering, TKM College of Engineering, Kollam
[4] Electronics & Communication Engineering, P.S.R. Engineering College, Sivakasi.
Tamil Nadu, India. Email: mebalu3@gmail.com

**Abstract**

Major cause of midvehicle collision is due to the distraction of drivers in both the Front and rear-end vehicle witnessed in dense traffic and high speed road conditions. In view of this scenario, a crash detection and collision avoidance algorithm coined as Midvehicle Collision Detection and Avoidance System (MCDAS) is proposed to evade the possible crash at both ends of the host vehicle. The method based upon Constant Velocity (CV) model specifically, addresses two scenarios, the first scenario encompasses two sub-scenario namely, a) A rear-end collision avoidance mechanism that accelerates the host vehicle under no front-end vehicle condition and b) Curvilinear motion based on front and host vehicle offset (position), whilst, the other scenario deals with parallel parking issues. The offset based curvilinear motion of the host vehicle plays a vital role in threat avoidance from the front-end vehicle. A desired curvilinear strategy on left and right sides is achieved by the host vehicle with concern of possible CV to avoid both end collisions. In this methodology, path constraint is applicable for both scenarios with required direction. Monte Carlo analysis of MCDAS covering vehicle kinematics demonstrated acute discrimination with consistent performance for the collision validated on simulated with real-time data.

**Keywords:** Midvehicle Collision Detection and Avoidance System, Constant Velocity, rear-end collision avoidance mechanism, offset based curvilinear motion, parallel parking

## 1 Introduction

Association for Safe International Road Travel (ASIRT) reveals approximately $518 billion dollar are expenditure every year globally. ASIRT computed most of the accidents are due to driver distraction, therefore Collision Detection and Avoidance Systems (CDAS) plays a vital role in automotive domain. According to National Transportation Safety Board (NTSB) report [1] demonstrates collision orient to rear-end and front-end of the vehicle on the highway road. Generally, such collisions are averted by estimating the target's co-ordinates and advancing the

host vehicle in the constraint path. In real time scenario, this demands fast computation algorithms for resolving such collisions. In [2], an adaptive control was presented that overtook the vehicle to avoid threat from the front-end, by considering the relative velocity between the two vehicles. The desired path for vehicle's further progressing was zero by determining the destination point of constraint path either off-road or on-road. Also, the size of the front-end vehicle plays a vital role in constructing the desired path [3]. Further improvements to lane switching mechanism was rendered in [4] [5] by blending GPS technology in lane detection. Although, effective this means was suitable only for low speed condition, as the base stations computational load, desired scaling to compute the accurate position of the vehicle in addendum to the intended curvilinear path required for lane switching.

Also, more coefficients and higher degree polynomials [6] are required to attain a smooth interpolated curvilinear path. Moreover, path estimation neglected the offset variations of the front-end vehicle position that resulted in a false constraint path to the host. Thereby, interpolating the desired path without offset region will threat the front-end vehicle. Computing curvilinear motion through imaging means was presented in [7]. This scheme suffered majorly from object and road cornering error that mislead to the collision. Moreover, the image-based motion control requires filter to reduce the aforementioned error that dramatically increased the system's complexity and was also cost ineffective. Rear-end collision avoidance using intercommunication amongst both end vehicles was realized in [8] [9]. This scheme was worthy for low speed conditions and non-synchronization between both the ends mislead and causes distraction to the driver-end. In high speed conditions, automatic guidance is preferred to avoid collision at both ends of the host vehicle. The vision-based method requires large memory and higher processing speed for computing the result of image. Moreover, vision-based methods are under research to reduce computational complexity.

One such backdrop is the host (Mid-vehicle) maneuvering in between the front and rear vehicle. Our intention is to deliver a novel MCDAS algorithm for intelligent tracking under inexorable condition to overcome both-end collisions. MCDAS deals with adaptive longitudinal and lateral (curvilinear) path estimation of host vehicle to sidestep collisions with adjoining vehicles. In contrary, existing methods rendering simple curvilinear motion crashes with the terminal region on road side or edges of front vehicle. The three fold contributions of the MCDAS are:

- Longitudinal path estimation of host vehicle depending on distance and velocity of the front and rear vehicles.

- Fusing suitable curvilinear path estimation under critical condition enforced by the front and rear vehicles.
- Additionally, path estimation scheme is included for parallel parking scenario.

## 2 Intended Methodology

Short-range is being more prone to collision incidents, an important fact that signifies the demand for development of short range applications in automotive technology particularly ensuring passenger's safety and comfort. Uncertain collision happens during congestion due to ambiguous drivers input in short range either in traffic or unblock condition. Focusing on evasive maneuvering of both (front and rear) vehicles and estimating an appropriate trajectory of the host vehicle in imminent situations is the novelty addressed in this paper. Detailed discussions of the intended model with relevant scenarios are dealt in subsequent sections.

*2.1 MCDAS*

Several kinematic parameters remain the backbone in formulating this model presented in Figure 1 and their behaviour was mainly considered in modelling this scheme.

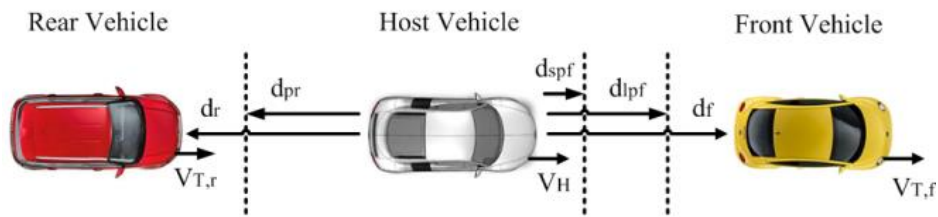

**Figure 1.** Midvehicle collision avoidance parameters

Thekinematic parameters in Figure 1 illustrates, $V_H$ - velocity of Host vehicle, $d_r$ - distance between Rear and Host vehicle, $d_{pr}$ - Predefined distances between Rear and Host vehicle, $V_{T,r}$ - velocity of Rear vehicle, $d_f$ - distance between Front vehicle and Host vehicle, $V_{T,f}$ - velocity of Front vehicle, $d_{lpf}$ and $d_{spf}$ are predefined range between Front and Host vehicle. Regulations for the identified parameters and their consequences summarized in Table 1 are considered by MCDAS to address the different crash conditions presented in Figure 2 and 3. These include No Front-End Vehicle in Figure 2 and Front-End Vehicle with an offset in Figure 3.

**TABLE 1**   Linguistic table proposed for MCDAS

| Model | FV | FV distance, $d_f$ [m] | FV velocity, $V_{T,f}$ [Km/h] | RV | RV distance, $d_r$ [m] | RV velocity, $V_{T,r}$ [Km/h] | Mode |
|---|---|---|---|---|---|---|---|
| MCDAS | No | $> d_{lpf}$ | None | Yes | $> d_{pr}$ | None | No acceleration |
|  | No | $> d_{lpf}$ | None | Yes | $\leq d_{pr}$ | $> V_H$ | Acceleration |
|  | Yes | $= d_{lpf}$ | $= V_H$ | Yes | $= d_{pr}$ | $= V_H$ | Warning |
|  | Yes | Between $d_{lpf}$ & $d_{spf}$ | $\leq V_H$ | Yes | $< d_{pr}$ | $> V_H$ | CCM (Forward direction) |
|  | None | None | None | None | None | None | CCM (Reverse direction) |

FV - Front Vehicle, RV - Rear Vehicle, CCM - Constraint Curvilinear Motion, $V_H$ - Host Vehicle Velocity, None -Unknown Value, $d_{pr} = d_{spf} = 25m$, $d_{lpf} = 35m$

Accordingly, predefined front and rear distance parameters for short range applications were adopted from [14], and the same values are reprised here in modelling MCDAS for the diverse crash scenarios.

*2.1.1   No Front-End vehicle*

This condition deals with a threat issued from the rear vehicle due to driver's inattention that may lead to a possible imminent collision with the host vehicle. To sidestep such imminent collision the host vehicle should maintain the appropriate velocity and distance with regard to the rear vehicle as shown in Figure 2. Based on $d_r$ and $V_{T,r}$ of the rear vehicle, the host vehicle is either suggested to advance further or proceed with the same acceleration. This scenario is mathematically framed in the first two rows of Table 1 based on the values of $d_{pr}, d_{lpf}$, in accordance with $d_f$ and $d_r$. Despite being a simple scenario, this was mainly considered as the majority of collisions happening at rear-end ascontinuous monitoring by the front vehicle driver is not possible at all times.

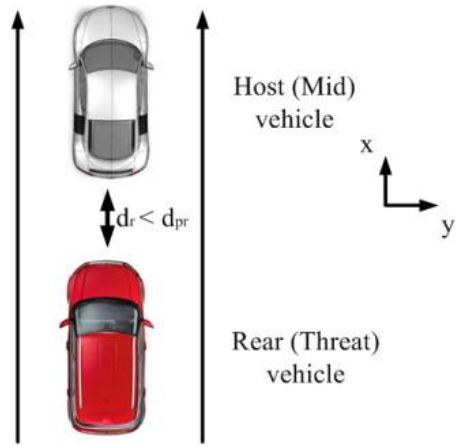

**Figure 2.** Imminent Collision avoidance under no front vehicle condition

*2.1.2 Front-End Vehicle with Offset-Based*

Experiencing the aforesaid road scene is not possible at all time in real road state, especially when roads are fully occupied. Therefore, front vehicle is also considered in this second condition, where host vehicle sandwiches the front and rear vehicles especially. During such crucial situations, the host vehicle has to acquire the prior trajectory condition of the scenario and accordingly maneuver to avoid collision. To address such road scenes, the values of $d_f$, $d_r$ and $V_H$ are estimated that are mapped respectively against $d_{lpf}, d_{pr}, V_{T,f}$ and $V_{T,r}$ and if found equal MCDAS warns the host vehicle to maintain constant speed. In another situation, where the distance parameters $d_f$ and $d_r$ are constrained between $d_{lpf}$, $d_{spf}$ and $d_{pr}$ with $V_H$ bounded by $V_{T,f}$ and $V_{T,r}$. Accordingly, the host vehicle must adapt a strategy path to avoid imminent collision at rear-end of host vehicle when the rear vehicle drives at harsh condition as illustrated in Figure 3.

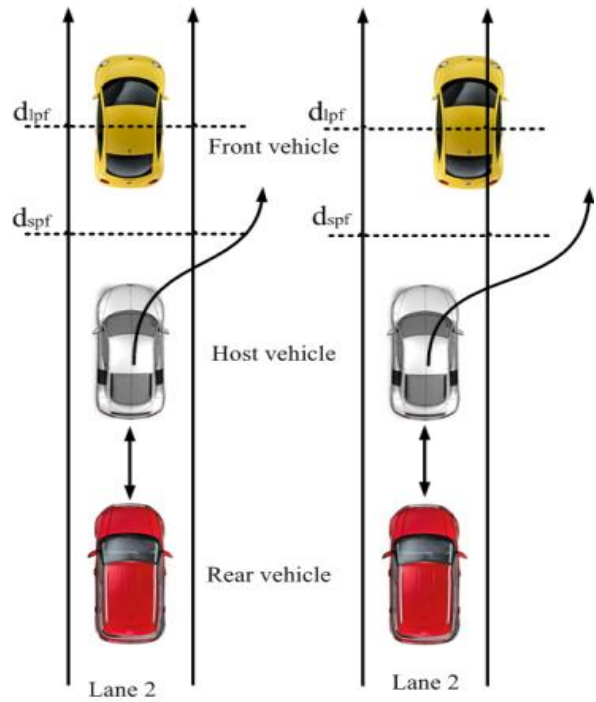

**Figure 3.** Curvilinear strategy of host vehicle based on front and host vehicle position

Mathematical conditioning of the aforesaid statements is formulated in Table 1. The front and host vehicle offsets(positions) are majorly employed for modelling an adaptive curvilinear motion. Offset-based lateral strategy is enclosed in the proposed MCDAS to avoid collision during curvilinear motion of host vehicle. Thus, the proposed method provides an intelligent trajectory to avoid mid-vehicle collision under imminent condition. Also, warning information is issued to the driver prior to the dead zone; if unattended later a smart constraint path is autonomously adapted by the host vehicle leading into the safe zone.

*2.1.3 Parallel Parking scenario*

Additionally, a parallel parking *scenario* is presented in this approach to address the parking issues faced by drivers that involves identification of constrained path in reverse direction. Moreover, the edges of the parked vehicle located at a certain destination point are majorly considered in parallel parking situations that demand optimal interpolation scheme for constraining the host vehicle's motion. In this regard, curvilinear motion along the reverse direction is suitable for Parallel parking mode as depicted in Figure 4. In this mode, the x and y-distances of the parked vehicle are fetched from the host vehicle's sensors (ESR). Depending on the maximum x and y-distance the comfortable curvilinear motion is interpolated without crashing the parked vehicles.

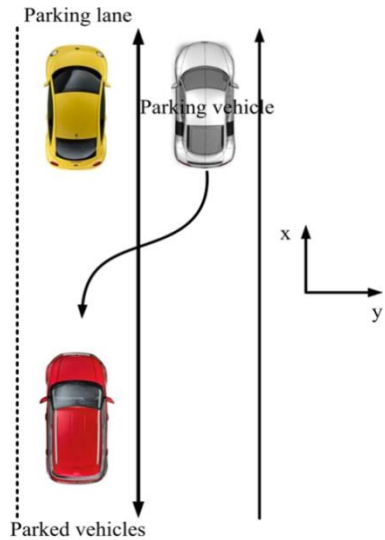

**Figure 4.** Reverse Curvilinear Trajectory model of parking vehicle

## 3  Modelling of MCDAS Model

Imminent collision at both ends of midvehicle can be evaded by possible intelligent motion model. MCDAS commences the path modelling operation by acquiring the current status of the host vehicle through the ESR sensors and accordingly decides to proceed with or without constraint curvilinear motion based on the crucial road conditions. Upon acquiring the self and surrounding kinematics information, MCDAS progresses for further action conditionally by identifying the presence of the front vehicle in the trajectory. Under no front vehicle condition, the mode selector commands adaptive speed control to maintain the host at desired speed with respect to rear vehicle. In another scenario, when the path is obstructed by both the front and rear vehicles in the forward and backward direction respectively, the kalman filter estimates (front, rear and host) the vehicles position at different instant of time and commands   mode selector command speed and trajectory generator to fit an appropriate constraint path with desired speed (collision free speed of host vehicle) before the crash zone to avoid imminent collision. An outline of the MCDAS is portrayed in Figure 5 that addresses the aforesaid concerns and the diverse road conditions.

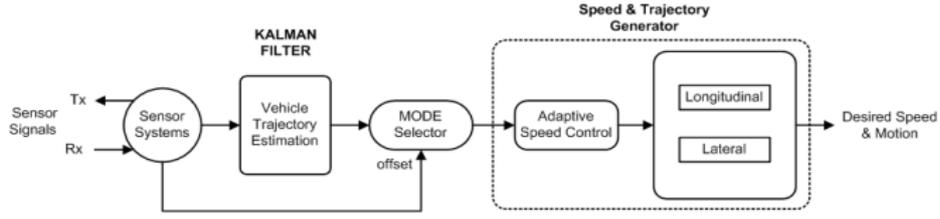

**Figure 5.** Trajectory model of MCDAS for different road scenarios

Systematic explanation of the individual modules and their course of action are detailed in relevant subsections.

*3.1 Kinematic Parameter Estimation*

At the outset, this block estimates the kinematic parameters of the three vehicles (Front, Host (Mid) and Rear) that corresponds to range, x-coordinate (x-position), y-coordinate (y-position), Doppler frequency ($f_D$) and relative velocity ($V_R$) acquired from the relevant measurement sensors the details of which are dealt in threat analysis section. The Range (R) amongst the vehicles is computed based on speed of light/sound (c) and delay time ($\tau$) in equation (1).

$$R = \frac{c\tau}{2} \quad (1)$$

Considering the curvilinear nature of MCDAS for path identification, the relevant position and distance parameters of the vehicles in scenario are calculated using the Euclidean distance metric. The x, y-positions of target (front and rear) vehicles are measured using trigonometric functions defined in equations (2) and (3).

$$x = R \cos \theta \quad (2)$$

$$y = R \sin \theta \quad (3)$$

Later, the Euclidean distance between two vehicles is obtained by target coordinates (x, y) and sensor coordinates ($x_s, y_s$) of host vehicle as shown in the equation (4).

$$d = \sqrt{(x - x_s)^2 + (y - y_s)^2} \quad (4)$$

Further refinement is contributed by Doppler frequency ($f_D$), speed of light (c), carrier frequency ($f_c$), received frequency ($f_r$) and transmitted frequency ($f_t$) in path estimation. The unknown target velocity ($V_T$) relies heavily upon on the host ($V_H$) and relative velocity ($V_R$) that is subsequently utilized for target path estimation. Target velocity is manipulated by considering target's x, y coordinates and velocity at different time instances in polar co-ordinatesas per Figure 6, where θ being target vehicle angle.

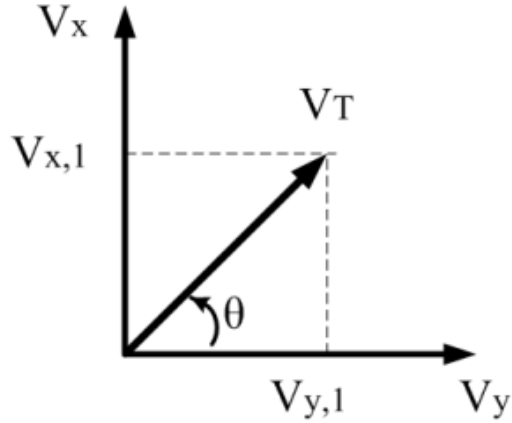

**Figure 6.** Phasor representation of target vehicle

Consequently, the above statements are mathematically modelled in equations (5 to 9).

$$V_R = \frac{cf_D}{2f_c} \quad (5)$$

$$f_D = f_r - f_t \quad (6)$$

$$V_T = V_H + V_R \quad (7)$$

$$v_x = V_T \sin\theta \quad (8)$$

$$v_y = V_T \cos\theta \quad (9)$$

Where, $V_T$ is addressed by $V_{T,f}$ (velocity of front vehicle) and $V_{T,r}$ (velocity of rear vehicle). The mathematical model directs MCDAS to proceed, either by longitudinal motion or adapt curvilinear motion to fix the constraint path. In this context the mitigation activities adopted by MCDAS under the above concerns are discussed in subsequent sections.

*3.1.1 Intelligent Longitudinal Motion of host vehicle*

An adaptive speed control system maintains a sufficient distance between two vehicles in order to sidestep crucial instants. Based on the predefined distance between rear (threat) and host vehicle the cruise control system will automatically accelerate the host vehicle in proportion to rear vehicle to avoid imminent collision. This scenario is earlier portrayed in Figure 2, where the host vehicle is adhered to maintain appropriate distance with rear vehicle. This mechanism is ensured by the cruise control system in automotive vehicles by employing Drive by Wire technology. A particular scenario addressed by the proposed MCDAS is explained below that considers the relative velocity of both rear, host vehicles at 60km/h and 40km/h respectively.

Based upon the acquired $V_{T,r}$ and subsequent evaluation of $D_{R-1}$ through equation (10), MCDAS then literally suggests the host vehicle to maintain a safer relative distance with the rear vehicle by appropriately varying the duty cycle, $D_H$ modelled in equation (11) of the drive by wire system thereby, avoiding imminent collision.

$$D_{R-1} = \frac{V_{T,r}}{V_{HM}} \tag{10}$$

$$D_H = D_{H-1} + \left(\frac{V_{T,r}}{V_{HM}} - D_{H-1}\right) \tag{11}$$

The assessed duty cycle for the parameters mentioned above is shown in Table 2.

**Table 2**      Duty cycle of host vehicle to avoid imminent collision with threat vehicle

| $D_{R-1}$ | $D_{H-1}$ | $V_{T,r}$ [Km/h] | $V_{HM}$ [Km/h] | $D_H$ |
|---|---|---|---|---|
| 0.25 | 0.16666667 | 60 | 240 | 0.25 |

$D_{R-1}$ - Duty Cycle of Rear Vehicle at previous state, $V_{HM}$ - Maximum Host Vehicle Velocity, $D_{H-1}$ - Duty Cycle of Host Vehicle at previous state.

Although longitudinal analysis resolves the rear-end vehicle collision in the absence of front vehicle, still, it fails to address situations, particularly, when the host vehicle is sandwiched between the front and rear. Under such constraints, lateral motion offers solution in the way of establishing a relative curvilinear path. To accomplish this under running conditions Kalman filter is incorporated into MCDAS for relative trajectory estimation.

*3.1.2 Kalman filter*

Rear and front vehicle trajectories are tracked at regular time intervals using kalman filter. The estimated position of front, host and rear vehicle assists in directing a curvilinear path before the collision point. In accordance with the acquired front and rear end vehicle velocities, the host vehicle is maneuvered at collision-free speed. Estimation begins by considering the front vehicle kinematics as the initial tracking point by the host vehicle and later path fitting is relatively modelled in accordance with the rear vehicle. MCDAS of the host vehicle is formulated using the CVmodel that tracks the linear kinematics of the front and rear vehicles. The mechanism commences with the state estimation of front and rear vehicle trajectories with process ($W_t$) and measurement ($V_t$) noise at continuous time interval using the Kalman models [14] presented below.

$$\begin{bmatrix} \hat{x}_r \\ \hat{\dot{x}}_r \\ \hat{y}_r \\ \hat{\dot{y}}_r \\ \hat{x}_f \\ \hat{\dot{x}}_f \\ \hat{y}_f \\ \hat{\dot{y}}_f \end{bmatrix} = \begin{bmatrix} 1 & T & 0 & 0 & 0 & 0 & 0 & 0 \\ 0 & 1 & 0 & 0 & 0 & 0 & 0 & 0 \\ 0 & 0 & 1 & T & 0 & 0 & 0 & 0 \\ 0 & 0 & 0 & 1 & 0 & 0 & 0 & 0 \\ 0 & 0 & 0 & 0 & 1 & T & 0 & 0 \\ 0 & 0 & 0 & 0 & 0 & 1 & 0 & 0 \\ 0 & 0 & 0 & 0 & 0 & 0 & 1 & T \\ 0 & 0 & 0 & 0 & 0 & 0 & 0 & 1 \end{bmatrix} \begin{bmatrix} x_r \\ \dot{x}_r \\ y_r \\ \dot{y}_r \\ x_f \\ \dot{x}_f \\ y_f \\ \dot{y}_f \end{bmatrix} + W_t \qquad (12)$$

$$\begin{bmatrix} \hat{x}_r \\ \hat{y}_r \\ \hat{x}_f \\ \hat{y}_f \end{bmatrix} = \begin{bmatrix} 1 & 0 & 0 & 0 & 0 & 0 & 0 & 0 \\ 0 & 0 & 1 & 0 & 0 & 0 & 0 & 0 \\ 0 & 0 & 0 & 0 & 1 & 0 & 0 & 0 \\ 0 & 0 & 0 & 0 & 0 & 0 & 1 & 0 \end{bmatrix} \begin{bmatrix} \hat{x}_r \\ \hat{\dot{x}}_r \\ \hat{y}_r \\ \hat{\dot{y}}_r \\ \hat{x}_f \\ \hat{\dot{x}}_f \\ \hat{y}_f \\ \hat{\dot{y}}_f \end{bmatrix} + V_t \qquad (13)$$

In the prediction phase, the position co-ordinates and the corresponding error variance are determined through equations (14) and (15). This error information represents the position deviations of the front and rear vehicles. $Q_{W_t}$, indicates system noise covariance matrix with zero mean. The computed kalman gain further assists in minimizing the variance of the vehicle position, thereby, presents acute trajectory information to the host.

$$\hat{X}_{t|t-1} = A\ \hat{X}_{t-1|t-1} \qquad (14)$$

$$P_{t|t-1} = A\ P_{t-1|t-1}A^H + Q_{W_t} \qquad (15)$$

This modification minimizes the Mean Square Error (MSE) that enhances the estimation process and establishes the true target position across different states realized through equations (16), (17) and (18). $R_{V_t}$ indicates measurement noise covariance matrix with zero mean.

$$K_t = P_{t|t-1}C^H[C\ P_{t|t-1}C^H + R_{V_t}]^{-1} \qquad (16)$$

$$\hat{X}_{t|t} = \hat{X}_{t|t-1} + K_t[Z_t - C\ \hat{X}_{t|t-1}] \qquad (17)$$

$$P_{t|t} = [I - K_t\ C]\ P_{t|t-1} \qquad (18)$$

For recursive 'n' state estimation, the initial state vector $\hat{X}_{t-1|t-1}$ is initialized to the posterior state vector $\hat{X}_{t|t}$ given in Equation (19).

$$\hat{X}_{t|t-1} = \hat{X}_{t|t} \qquad (19)$$

Vehicle trajectories estimated by the proposed MCDAS for diverse road scenarios are presented in Figure7,8 and 9.The strength of MCDAS for addressing diverse road crash scenarios is better understood using Monte-Carlo simulations.

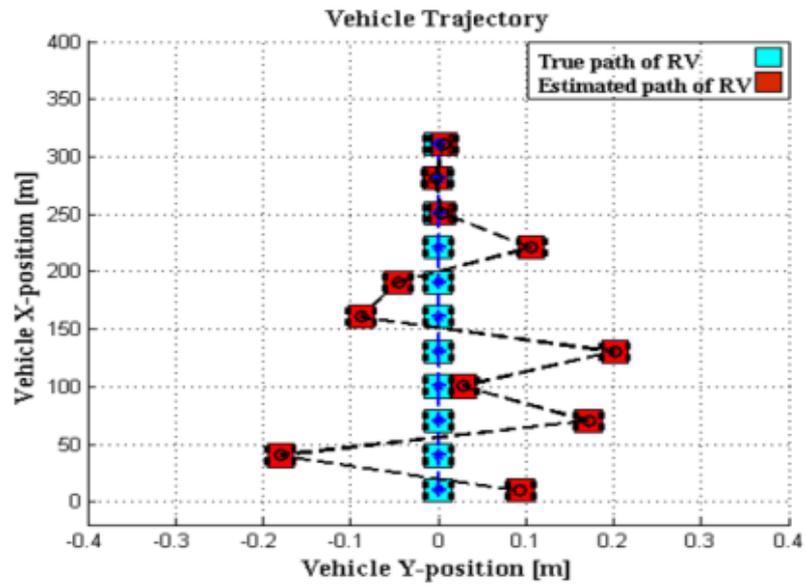

**Figure 7.** Estimated trajectory of RV-Rear Vehicle (red patch)

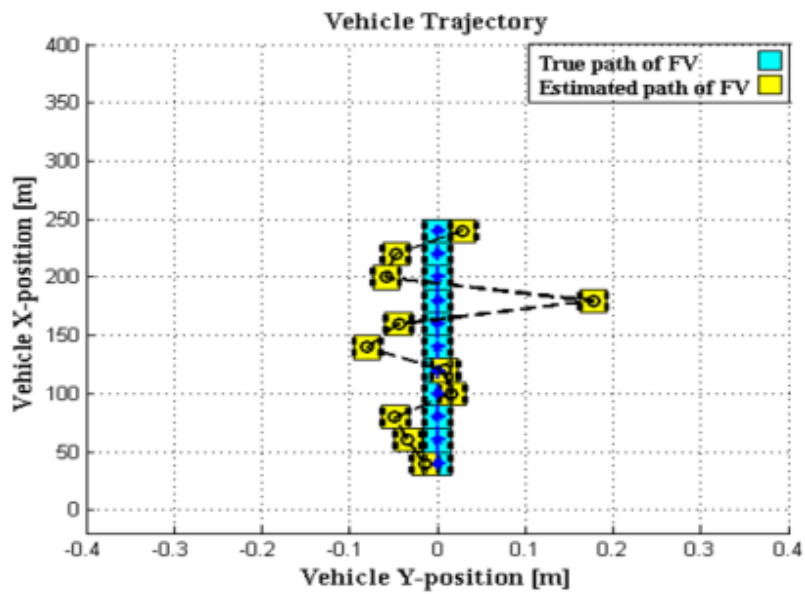

**Figure 8.** Estimated trajectory of FV-Front Vehicle (yellow patch)

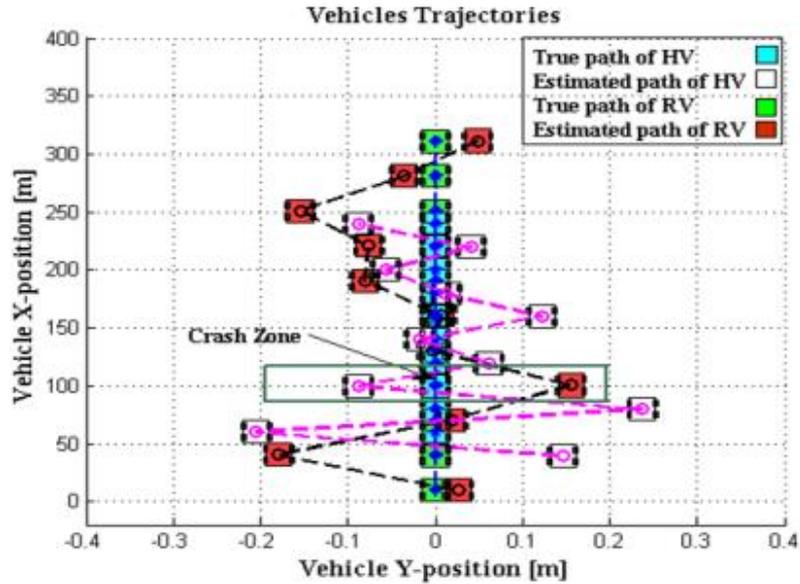

**Figure 9.** Estimated trajectory of HV-Host Vehicle (white patch) and prediction of crash zone between host and rear vehicle

*3.1.3 Threat Analysis Using Monte Carlo Simulation*

The MCDAS efficacy is studied using offline simulations on vehicle kinematic data acquired from Electronically Scanning Radar (ESR)detailed in [15] of Delphi automotive systems under two scenarios namely midvehicle collision and parallel parking covered by MCDAS. The diverse trajectory estimates include: Figure 7 deviations between the true and estimated path of the rear with regard to host. Likewise, Figure 8 depicts the path deviations of front with respect to host and finally, Figure 9 illustrates variations of the true and estimated paths of rear and host vehicles and its collision zone.

The trajectories established in Figure 7, 8 and 9 were evaluated within a time frame of 0.5 seconds. The true and estimated paths of the different vehicles are represented by unique colours to highlight the significance of MCDAS. In Figure 7 the true path is highlighted by the blue vehicle and the red vehicle that corresponds to the estimated path with a velocity of 60Km/h. The deviations between the estimated and the true path are constrained between -0.2 to 0.2 meters of the vehicle y-position at the commencement of the estimation phase and finally the gap closes at 250m along the x-position. Thus, MCDAS settles down instantaneously to the true path.

A similar scenario is witnessed in Figure 8 that corresponds to the trajectory estimation of the front vehicle travelling with a velocity of 40Km/h by the MCDAS. As mentioned above MCDAS settles down swiftly thus, minimizing the risks of collision. In this particular scenario, the estimated trajectory of both vehicles characterizes the amount of distance travelled over the time span which is around 6 seconds (11 samples with 0.5 seconds each). Under such vehicle constraints, the front vehicle catches the true path at 250meter from the initial position with minimal deviations evident in Figure 8.

Moreover, in this proposed method the front, rear and host vehicles are travelling at 40, 60 and 40km/h respectively. Accordingly, MCDAS effectively predicts crash zone for this scenario as observed in Figure 9. The front vehicle trajectory is not predicted in Figure 9 due to same velocity with respect to host vehicle. The suitable trajectory is need to be estimated for host vehicle in order to take a constraint path before the crucial zone (100-120m) at which an imminent collision may occur as illustrated in Figure 9. Overall from the above cases it is evident that the error estimates of MCDAS are highly constrained and achieves rapid convergence thereby, offering amicable solution to road problems encountered during longitudinal motion.

*3.1.4 Lateral Motion*

The identification of crash zone in the aforesaid scenario further pushes MCDAS for resolving vehicle crashes using lateral motion. In particular situations, wherein the host vehicle is sandwiched between the front and rear, it is desired that a relative velocity has to be maintained amongst them. Complex road scenarios such as the above will be effectively addressed by the proposed MCDAS that constructs a constraint curvilinear path prior to the crash spot that avoids imminent collision. Moreover, a predefined path fitting the above road scenario, with fixed offsets does not suit in real time due to rigorous road conditions as illustrated in Figure 10.

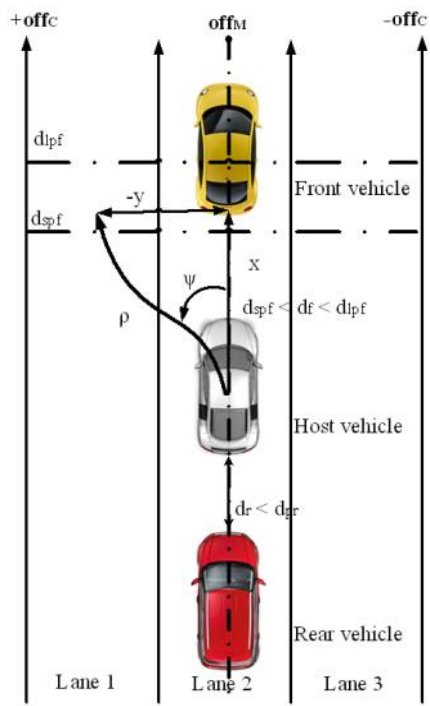

**Figure 10.** Fixed lateral motion of host vehicle

This constraint is also addressed by MCDAS owing to the integrated blind curvilinear motion estimation scheme. Neglecting $off_F$ and $off_H$ knowledge in real road condition will lead to threat against the vehicles. Thereby, $off_F$ and $off_H$ play a major role in determining the constraint curvilinear motion either on the left or right side of the host. Thus, offset-based blind curvilinear estimation is modelled in MCDAS.

*3.1.5 Lateral Motion with Offset*

$off_H$ and $off_F$ are essential for deciding the offset (y-position) that is added to the constraint path for avoiding crash on front vehicle edges and establish an adaptive curvilinear trajectory for host vehicle is illustrated in Figure 11.

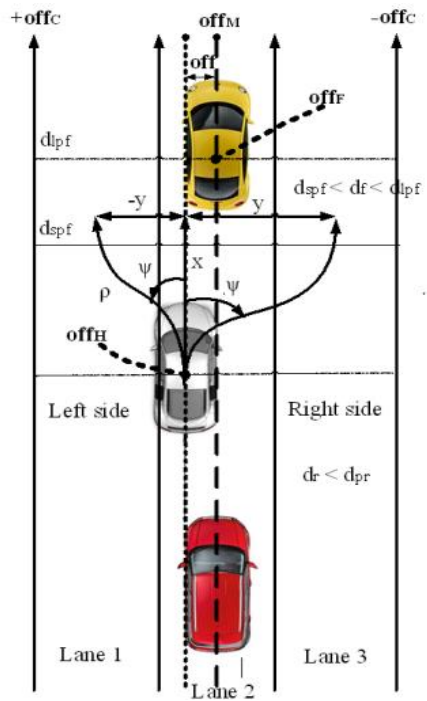

**Figure 11.** Proposed suitable curvilinear motion

The lateral distance between the host and front vehicle describes the offset (off) to be incorporated as given below.

$$\text{off} = \text{off}_F - \text{off}_H \quad (20)$$

Practically, the proposed equation (20) manipulates adaptive motion without threating the front vehicle. Offset value takes less or more amount of y-distance on Left side or Right side depending on the front vehicle position through which the host vehicle fits appropriate curvilinear motion on LSD or RSD as illustrated in Figure 11. The parameters '–y' and 'y' signifies y-position of host vehicle on the LSD and RSD, where the corresponding angle and range being represented as $\psi$ and $\rho$ respectively.

The curvilinear equations framed in (21), (22) and (23) calculate the position of host vehicle as a function of time.

$$x = K\,t \tag{21}$$

$$y = \frac{y_{max} + off}{1 + be^{a x_t}} - y(0)\ ; \{Right\} \tag{22}$$

$$y = -\frac{y_{max} - off}{1 + be^{a x_t}} + y(0)\ ; \{Left\} \tag{23}$$

The constant parameter K, establishes suitable sample point of x-position of host vehicle. The off parameter represented in equations (22) and (23) are modeled along with coefficients and dependent variable in a sigmoidal fashion to configure an adaptive curvilinear motion. Additionally, the parameter y(0) incorporated in (22) and (23) nullifies at time t = 0. Moreover, trajectory constants a, b and $y_{max}$ fused in y-position to assists convenient trajectory in (22) and (23). Therefore, the above position equations are suitably interpolated for intelligent trajectory estimation. The equations (21) and (22) are interpolated at respective time instant for right side curvilinear trajectory of host vehicle. Similarly, equations (21) and (23) are interpolated at respective time instant for left side curvilinear trajectory of host vehicle.

The kinematic behaviors of the modelled equations are illustrated in the following figures that represent a generic road scenario comprising of three lanes. An investigation of the proposed MCDAS with different offsets under vehicle running conditions is dealt here.

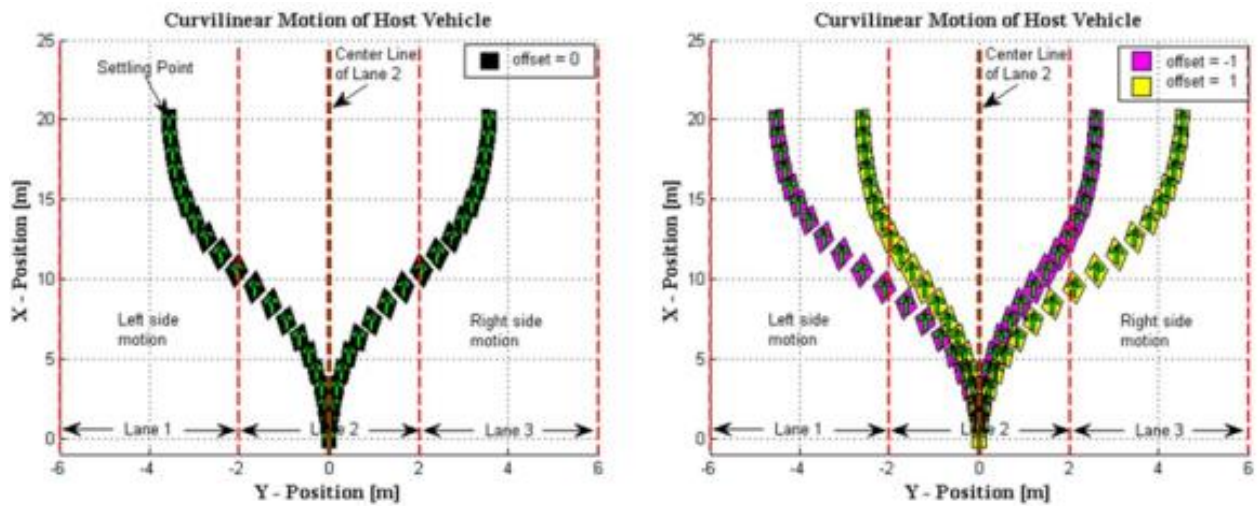

**Figure12(a) and 12(b).** Estimated trajectories of host vehicle for 12(a). offset = 0 and 12(b). offsets = (-1 and 1)

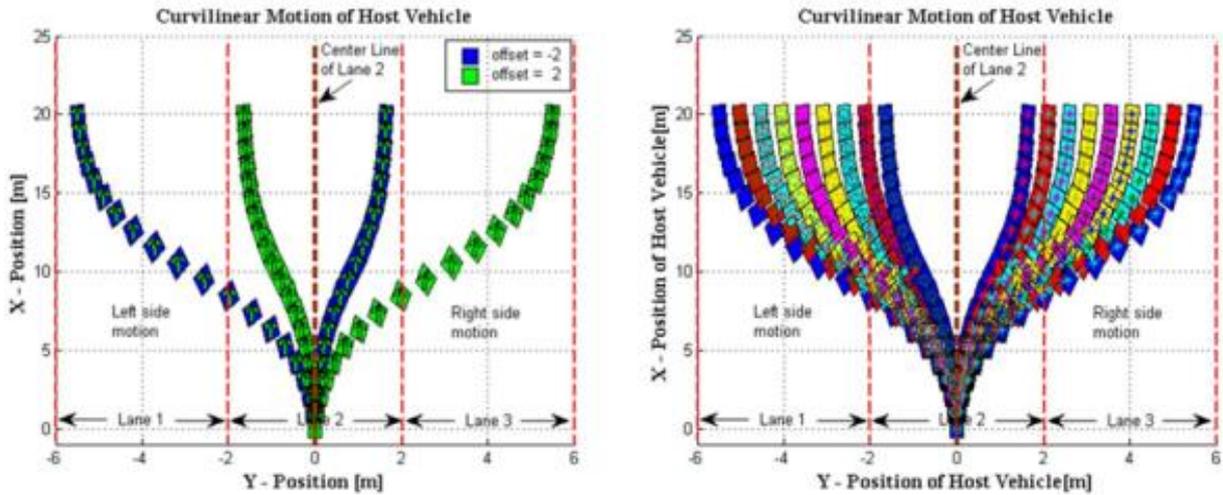

**Figure 12(c) and 12(d).** Estimated trajectories of host vehicle for 12(c). offset = (-2 and 2)and 12(d). offsets = (-2, -1.5, -1, -0.5, 0, 0.5, 1, 1.5, 2)

In the above figures the possible curvilinear trajectories that can be adhered by the vehicles for diverse conditions is presented. Figure 12(a) reveals the vehicles curvilinear trajectories under zero offset condition. The offset based on the front and host vehicle provide suitable motions,thereby non-threating the front vehicle. Furthermore, depending on the offset values of the front and host vehicle flexible motion is obtained by host vehicle, as shown in Figure 12(b) and 12(c). If the offset value is positive, the estimated path is more diverged along the right hand with lesser deviation along the left hand side with regard to the appropriate lanes. In contrast if the offset value is negative the model shifts its trajectory towards the left hand side extending to lane1 from lane 2. The Figure 12(d) highlights trajectory information of host vehicle for different offset values. Thus, a small change in the offset value is accordingly fitted by the MCDAS that assists in switching the lanes based upon the modelled trajectories thereby, avoiding collision.

To further examine the efficacy of MCDAS in estimating the curvilinear trajectories, error analysis is performed. This trajectory error is determined by measuring the distance between the host and the front vehicle after reaching the settling point determined by the curvilinear trajectory along the y-direction. The mechanism commences with the nullification of initial parameters pertaining to x-position, 'y' along Left Side Direction (LSD) and 'y' along Right Side Direction (RSD) positions using y(0) determined by assuming t=0, $y_{max}$=3.7m which corresponds to the maximum separation between the center line of two lanes recursively using equations (22) and (23). This arrangement nullifies the initial position error at zeroth time instant while taking a curvilinear motion by host vehicle. Accordingly, the behaviour of MCDAS under diverse road conditions for different trajectories is analysed using the position errors evaluated and reported in Table 3.

**Table 3**     Initial and Final position error during estimated path of host vehicle

| Offset [m] | Initial Position | | | Final Position | | | | | | |
|---|---|---|---|---|---|---|---|---|---|---|
| | x [m] | y (LSD) [m] | y (RSD) [m] | x [m] | y (LSD) [m] | Absolute y (LSD) [m] | y-position Error [m] | y (RSD) [m] | Absolute y (RSD) [m] | y-position Error [m] |
| -2 | 0 | 0 | 0 | 20 | -5.4942056 | -5.7 | -0.2057944 | 1.6386227 | 1.7 | 0.0613773 |
| -1.5 | 0 | 0 | 0 | 20 | -5.0122578 | -5.2 | -0.1877422 | 2.1205706 | 2.2 | 0.0794294 |
| -1 | 0 | 0 | 0 | 20 | -4.5303099 | -4.7 | -0.1696901 | 2.6025185 | 2.7 | 0.0974815 |
| -0.5 | 0 | 0 | 0 | 20 | -4.048362 | -4.2 | -0.151638 | 3.0844663 | 3.2 | 0.1155337 |
| 0 | 0 | 0 | 0 | 20 | -3.5664142 | -3.7 | -0.1335858 | 3.5664142 | 3.7 | 0.1335858 |
| 0.5 | 0 | 0 | 0 | 20 | -3.0844663 | -3.2 | -0.1155337 | 4.048362 | 4.2 | 0.151638 |
| 1 | 0 | 0 | 0 | 20 | -2.6025185 | -2.7 | -0.0974815 | 4.5303099 | 4.7 | 0.1696901 |
| 1.5 | 0 | 0 | 0 | 20 | -2.1205706 | -2.2 | -0.0794294 | 5.0122578 | 5.2 | 0.1877422 |
| 2 | 0 | 0 | 0 | 20 | -1.6386227 | -1.7 | -0.0613773 | 5.4942056 | 5.7 | 0.2057944 |

It can be witnessed from Table 3, the three initial position columns are packed with zeros against the different offsets an attribute that is owed to y(0). Further, error analysis is performed upon reaching the settling point by the host vehicle at final time instant as highlighted in Figure 12(a). In accordance with equation (21), absolute x-position at the settling point is assumed to be 20m that corresponds to zero x-position error. Similarly, the y-position deviations reported in Table 3 between the observed and absolute y-position is stated in the respective position error columns for the LSD (Left side) and RSD (Right side) parameters with respect to the different offsets. For a particular offset value of -2, the y-position error is determined to 0.2057944 on LSD and 0.0613773 on RSD respectively. Similarly, y-position error is evaluated for -1.5, -1, -0.5, 0, 0.5, 1, 1.5, 2 offset values on LSD and RSD which clearly indicates the potential of MCDAS that offers rapid convergence and higher adaptive to the rigorous road conditions. Moreover, the maximum deviation bounds of the curvilinear trajectory is evaluated and presented in Table 4.

**Table 4**    y-position Error Range at settling point for various offset values

| Offset [m] | $y_{max}$ [m] | Coefficients | | y-position Error Range (LSD) [m] | y-position Error Range (RSD) [m] |
|---|---|---|---|---|---|
| | | a | b | | |
| All values | 3.7 | -0.4 | 50 | -0.0613773 to -0.2057944 | 0.0613773 to 0.2057944 |

It can be seen that the y-position error for different offset values is confined between -0.0638m to -0.2057944m for LSD of host vehicle and +0.0638m to +0.2057944m for RSD using the constants 'a' and 'b' as -0.4 and 50 to obtain suitable constraint path.

Prior to adopting the curvilinear motion and before settling at the crash zone the MCDAS examines the road scenario and accordingly prepares the host vehicle to suit the rigorous road condition using equation (24).

$$\begin{bmatrix} V_{x,H} \\ V_{y,H} \\ \psi_t \\ \rho_t \end{bmatrix} = \begin{bmatrix} \sin\left(\tan^{-1}\frac{x}{y}\right) & 0 & 0 & 0 \\ \cos\left(\tan^{-1}\frac{x}{y}\right) & 0 & 0 & 0 \\ 0 & 0 & \tan^{-1}\left(\frac{x}{y}\right) & 0 \\ 0 & 0 & 0 & \sqrt{x^2+y^2} \end{bmatrix} \begin{bmatrix} V_{HM}\left(D_{H-1} + \left(\frac{V_{T,r}}{V_{VM}} - D_{H-1}\right)\right) \\ 0 \\ 1 \\ 1 \end{bmatrix} \quad (24)$$

The kinematic model of equation (24), educates the host vehicle to fit the appropriate parameters into its maneuvering phase. The equation (24) represents the amount of velocity to be maintained by the host vehicle while taking a constraint path to avoid collision with rear vehicle by additionally considering the following kinematic parameters namely Rear Vehicle Velocity ($V_{T,r}$), Host Vehicle Velocity ($V_{VM}$), duty cycle ($D_{H-1}$) at '$t-1$' time and finally, the Maximum Host Vehicle Velocity $V_{HM}$. Similarly, the other essential parameters for ensuring the vehicle safety that are to be determined is presented in the left hand side vector of equation (24) and are labelled as Host Vehicle Velocity along x and y-axis respectively along with instantaneous angle and range of host vehicle.

*3.1.6 Parallel Parking kinematics*

The potential of MCDAS is further exploited by extending it to a parallel parking scenario due to its inherent constraint curvilinear path fitting mechanism. Parallel parking situation demand curvilinear path fitting strategy in the reverse direction by the vehicle, whilst, settling down at the specified point as shown in Figure 13. The ESR sensors estimate the maximum distance along 'x' (longitudinal) and 'y' (lateral) directions through which a predefined curvilinear path could be designed for this application.

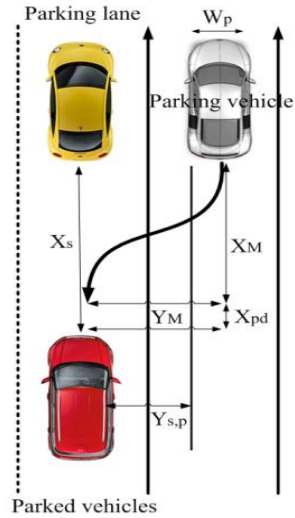

**Figure 13.** Parallel Parking strategy

The x-distance sensor (ESR) measures the amount of maximum distance ($x_s$) between the two parked vehicles. A predefined distance ($x_{pd}$) along x-direction between parking and parked vehicle is user defined based on subjective analysis. The maximum x-distance ($x_m$) in equation (25) is interpolated to fit the curvilinear path for the parking vehicle. Simultaneously, the y-distance sensor (ESR) provides two distance measurements one being the parked vehicle distance and another corresponds to the non-parked area distance present across the left-side lane. The maximum y-distance ($y_m$) for the parking vehicle in equation (26) is designed based on the parked vehicle distance ($y_{s,p}$) and half of the width of parking vehicle ($\frac{W_P}{2}$).

$$x_m = x_s - x_{pd} \qquad (25)$$

$$y_m = y_{s,p} + \frac{W_p}{2} \qquad (26)$$

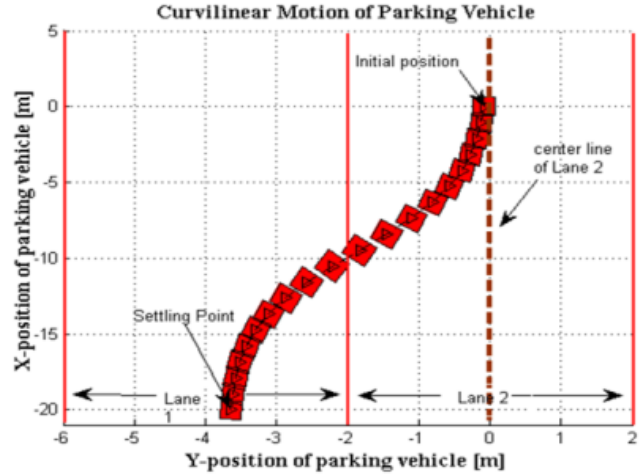

**Figure 14.** Simulated Curvilinear (reverse) motion of parking vehicle

To model the curvilinear motion, a sigmoidal function fits to parallel parking application. Figure 14 illustrates motion of parking vehicle in reverse direction with suitable interpolation of equations (27) and (28).

$$x = -K t \qquad (27)$$

$$y = \frac{-y_{max}}{1+be^{ax_t}} + y(0) \qquad (28)$$

Error analysis is also performed to investigate the efficacy of MCDAS for parallel parking schemes. The reverse path for the host vehicle is obtained with appropriate fitting coefficients using the values presented in Table 5. For the listed values the y-position error settles at 0.06 to 0.08 meter which indicates that beyond this range constructing the constraint path for the host vehicle is difficult in the parking scenario. In contrary, adopting inappropriate coefficient values ('a' and 'b') also, leads to inadequate curvilinear motion that does not suit parallel parking.

**Table 5** y-position error range for variable coefficient values

| Coefficients | | $y_{max}$ | y-position Error Range |
|---|---|---|---|
| a | B | [m] | [m] |
| 0.4 | 45 to 60 | 3.7 | 0.06 to 0.08 |

## 4  Conclusion

This paper describes two CV models for mid-vehicle collision detection and avoidance with parallel parking facility. The intended MCDAS avoids imminent collision by modelling various curvilinear motion based on the offset of the front and host vehicles. Although, curvilinear motion is widely utilized in several automotive applications the ability to deliver an efficient offset based-curvilinear motion by the host vehicle is still lacking, that is resolved in this proposal. A blind curvilinear motion is not prescribed at all-times due to real-time impact thereby; offset-based curvilinear motion is prescribed to avoid crash on or by the host vehicle.The settling position of host vehicle has position error ranging between -0.0613773 to -0.2057944 meter on LSD and 0.0613773 to 0.2057944 meter on RSD along y-distance is the limitation of the proposed method.  In parallel parking scheme, a constrained curvilinear motion (reverse direction) is interpolated with suitable coefficients to obtain a proper settling point. The settling position of parking vehicle after taking curvilinear motion has position error lying between 0.06m to 0.08m based on suitable coefficients. Exhaustive kinematic analysis of the modelled MCDAS reveals its ability to address the wider issues encountered by vehicles in diverse road conditions. Future research will look at extending this concept to intelligent vehicle transportation systems.